\documentclass[letterpaper]{article} 
\usepackage{aaai24}  
\usepackage{times}  
\usepackage{helvet}  
\usepackage{courier}  
\usepackage[hyphens]{url}  
\usepackage{graphicx} 
\usepackage{natbib}  
\usepackage{caption} 
\usepackage{algorithm}
\usepackage{algorithmic}
\usepackage{subcaption}
\usepackage{amsmath}
\usepackage{amssymb}
\usepackage{mathtools}
\usepackage{amsthm}
\usepackage{url}

\DeclareMathOperator{\Err}{Err} 
\DeclareMathOperator{\topt}{top}
\newcommand{\by}{\mathbf{y}}
\newcommand{\bx}{\mathbf{x}}

\theoremstyle{plain}
\newtheorem{theorem}{Theorem}

\newtheorem{problem}[theorem]{Problem}

\theoremstyle{definition}
\newtheorem{definition}[theorem]{Definition}

\theoremstyle{remark}
\newtheorem{remark}[theorem]{Remark}
\usepackage{dsfont}
\usepackage{newfloat}
\usepackage{listings}
\DeclareCaptionStyle{ruled}{labelfont=normalfont,labelsep=colon,strut=off} 
\floatstyle{ruled}
\newfloat{listing}{tb}{lst}{}
\floatname{listing}{Listing}
\title{Dissenting Explanations: \\ Leveraging Disagreement to Reduce Model Overreliance}
\author{
    Omer Reingold\equalcontrib,
    Judy Hanwen Shen\equalcontrib,
    Aditi Talati\equalcontrib
}
\affiliations{
    Stanford University\\

    \{reingold, atalati, jhshen\}@stanford.edu
}

\usepackage{bibentry}

\begin{document}

\maketitle

\begin{abstract}
While modern explanation methods have been shown to be inconsistent and contradictory, the explainability of black-box models nevertheless remains desirable. When the role of explanations extends from understanding models to aiding decision making, the semantics of explanations is not always fully understood – to what extent do explanations ``explain” a decision and to what extent do they merely advocate for a decision? Can we help humans gain insights from explanations accompanying \textit{correct} predictions and not over-rely on \textit{incorrect} predictions advocated for by explanations? With this perspective in mind, we introduce the notion of dissenting explanations: conflicting predictions with accompanying explanations. We first explore the advantage of dissenting explanations in the setting of model multiplicity, where multiple models with similar performance may have different predictions. Through a human study on the task of identifying deceptive reviews, we demonstrate that dissenting explanations reduce overreliance on model predictions, without reducing overall accuracy. Motivated by the utility of dissenting explanations we present both global and local methods for their generation.
\end{abstract}

\section{Introduction}
The development of increasingly capable AI systems has motivated the adoption of AI-assisted decision-making. In high-stakes settings such as loan approval and patient diagnosis, it is imperative for humans to understand how any given model came to its decision. However, with the success of deep learning, many large state-of-the-art models are not easily interpretable. Thus, explainability (XAI) methods are crucial for providing justification for the decisions of black-box models. Such explanations justify a model's prediction on a singular input example, and their goal is to provide accurate information while also being succinct and easy for humans to parse \cite{burkart2020survey}. 

\citet{wang2021explanations} list the desiderata of explanations as (1) model understanding, (2) helping people recognize model uncertainty, and (3) calibrating trust for AI models. Towards the goal of understanding a single model, recent works have shown that explanations generated from different methods based on the same instance can conflict \cite{han2022explanation, krishna2022disagreement}. While multiple explanations for the same \textit{model} question the validity of explainability tools, diverse and even conflicting explanations for the same \textit{decision} might better calibrate trust and illustrate uncertainty in a model-aided decision-making setting. 
Instead of rejecting explanations altogether, the existence of multiple plausible explanations motivates the perspective that explanations can be treated as arguments supporting a given model recommendation in the decision-making process. 

With the framework of explanations as arguments, we may naturally construct a courtroom analogy, in which human decision makers are the judges deciding whether the model prediction is trustworthy. When a singular explanation is provided, a decision-maker may be unduly influenced to trust the prediction. Indeed, \citet{bansal2021does} show that when explanations are provided, humans are more likely to follow a model decision regardless of whether the model is correct. Thus, while an explanation provides a supporting argument for a prediction, we must also provide alternative arguments, arguing against the model prediction, in order to accommodate meticulous human decision-making. In the context of a consequential legal decision, presenting both sides amounts to procedural due process.

In this paper, we introduce the notion of \textit{dissenting explanations}: explanations for an opposing model prediction to some reference model. To illustrate the importance of these explanations, we focus, from human study to proposed techniques, on a single task: deceptive hotel reviews classification. We perform a study to show that, on this difficult-to-verify task, dissenting explanations indeed reduce model overreliance without reducing the accuracy of the human predictions. Finally, since dissenting explanations are a useful tool for reducing overreliance, even outside the context of existing model multiplicity, we develop methods to induce predictive multiplicity and create dissenting explanations. We present techniques for generating global disagreement with respect to any black-box model, as well as local disagreement on any instance; these methods achieve disagreement without sacrificing model accuracy 

\section{Related Work}

\paragraph{One Model, Multiple Explanations}
Post-hoc explanations can be elicited from black box models through techniques including perturbation-based methods \cite{ribeiro2016should, lundberg2017unified} and gradient-based methods \cite{selvaraju2017grad, smilkov2017smoothgrad}. 
However, when applying such techniques to the same example, inconsistent and conflicting explanations for feature importance may arise. Surveying data scientists, \citet{krishna2022disagreement} found disagreements in feature explanations and find that more complex models exhibit higher disagreement.  

\paragraph{Similar Models, Conflicting Explanations}
For trustworthy predictions, humans may expect similar predictions from similarly accurate models. Yet models with similar accuracy still exhibit different predictions \cite{marx2020predictive}.  \citet{brunetimplications} show that similar accuracy models can yield vastly different explanations due to different random seeds and hyperparameters. Other works in robustness purposefully generate disagreement between models \cite{pang2019improving, rame2021dice}. 

\paragraph{Overreliance and Human-AI Collaboration}
Among tasks where neither humans nor AI routinely achieves perfect performance, \citet{lai2019human} use AI predictions and explanations to help human participants with detecting deceptive hotel reviews and find that human performance was improved with AI predictions with explanations. \citet{vasconcelos2022explanations} also find that explanations actually reduce overreliance in their set of maze task experiments. In contrast, \citet{bansal2021does} study common sense tasks including review sentiment classification and found that explanations increased accuracy when the AI model was correct but decreased accuracy when the AI model was wrong. An adjacent line of work studies AI debate: two large language models are prompted to recursively try to convince a human judge to take their side \cite{michael2023debate}. 

We investigate the effect of also showing the explanation of a dissenting model in reducing overreliance in settings where AI surpasses human performance, but human decision-makers may need to make the final decision \cite{lai2019human, lundberg2017unified}. In these settings, the goal is to provide AI predictions with explanations to humans as a tool rather than removing humans from decision-making altogether. The closest prior work to ours includes a position paper arguing for decision support systems to provide support for and against decisions \cite{miller2023explainable} and the effect of using different explanations from the same model \cite{bansal2021does}. In contrast, we examine differing independent predictions and accompanying explanations from different models in the Rashomon set \cite{fisher2019all} with the goal of improving human decision-making. 

\section{Model and Framework}
We define dissenting explanations in the situation where we have model multiplicity. Let $f, g : \mathcal{X} \rightarrow \mathcal{Y}$ be two different functions trained on the same data ${x, y} \sim \mathcal{D}$; these functions do not have to belong to the same hypothesis class. We look at the specific case of binary classification ($y \in \{0, 1\}$), but much of this work can also be extended to general classification tasks. 
Then, let $e(f, x)$ be an explanation for the model's prediction $f(x)$. The shape of $e$ depends on the type of explanation being used, and any of the standard explanation methods will produce a valid function $e$. Based on these definitions, we introduce the concept of a \textit{dissenting explanation} as an explanation of the prediction of a disagreeing model: 
\begin{definition}[Dissenting Explanation]
Let $f, g$ be any two different classifiers and let $(x, y) \sim \mathcal{D}$ be any example. Then, $e(x, g)$ is a \textit{dissenting explanation} for $e(x, f)$ if $f(x) \ne g(x)$. 
\end{definition}
Dissenting explanations offer an argument for a contradictory prediction; each disagreeing model can produce its own dissenting explanation. Furthermore, dissenting explanations are explanation-method agnostic. In the more general setting of multi-class classification, the explanation $e(g, x)$ is a dissenting explanation for $e(f,x)$ as long as $g$ predicts a label different from $f(x)$. 

Since disagreeing predictions are necessary for dissenting explanations, measuring how many predictions $f$ and $g$ disagree on gives an indication of how many dissenting explanations can be generated between two models. 
\begin{definition}[Global Predictive Disagreement]
Let $f, g$ be any two different classifiers, the global disagreement between $f$ and $g$ on some set $D$ is: 
\[
\delta_D(f, g) = \frac{1}{|D|}\sum_{x \in D}\mathds{1}[f(x) \ne g(x)]
\]
\label{def:global_disagree}
\end{definition}

\begin{remark}
Let $\Err_D(f) = \frac{1}{|D|}\sum_{(\bx, \by) \in D}\mathds{1}[f(x) \ne y]$ be the empirical error of a classifier. For two classifiers $f$ and $g$ where $\Err_D(f), \Err_D(g) \in [0, 1]$, then: 
\[
\delta_D(f, g) \le \Err_D(f) + \Err_D(g)
\]This can be seen by considering that disagreement is maximized when $f$ and $g$ make mistakes on disjoint sets. 
\label{rem:upper_bound}
\end{remark}

While disagreement can be maximized by models that disagree on every example, we care about the setting of models with similarly high accuracy, also described as the Rashomon set \cite{fisher2019all}.

Following prior work studying the overreliance of humans on AI predictions \cite{vasconcelos2022explanations}, we define overreliance as how much human decisions mirror AI suggestions when the AI is incorrect: ${\mathbb{E}[h(x) = f(x) | f(x) \ne y]}$ where $h$ represents the human decision. 

For the purposes of our experiments, we let $e(x, f) \in \mathbb{R}^d$ be a \textit{feature attribution explanation} of $f$ on $x$. A feature attribution explainer generates a linear ``surrogate model'' that approximates $f$ in a neighborhood of $x$. If the weights of the linear surrogate model are $w_i$, then $e(x, f)$ returns the most important features $x_i$, corresponding to the $d$ largest values of $|w_ix_i|$. 
For a feature attribution explanation, we denote ${e(x, f)_+ \in \mathbb{R}^p}$ as the set of features supporting the prediction $f(x)=1$ while $e(x, f)_- \in \mathbb{R}^n$ are the set of features that support the prediction $f(x)=0$ where $p+n=d$.

\section{Motivating Study: The Importance of Dissenting Explanations}
\paragraph{Hypothesis}
Motivated by the potential of dissenting explanations to present an alternative argument against a model prediction, we seek to understand whether dissenting explanations can be helpful in reducing human overreliance on model predictions. We propose two hypotheses:
\begin{quote}
\textsc{Hypothesis 1} (\textbf{H1}): Providing users with a singular explanation for an incorrect AI prediction increases human agreement with the incorrect prediction.

\textsc{Hypothesis 2} (\textbf{H2}): Providing users with a dissenting explanation, arguing against the AI prediction, along with the explanation, will decrease human over-reliance without significantly decreasing human accuracy, as compared to providing a single AI prediction and explanation.
\end{quote}

The purpose of the first hypothesis was to provide a baseline for how explanations affect human decisions, while the second hypothesis tests the value of dissenting explanations. We limit the scope of these hypotheses to a specific task which we investigate thoroughly. 

\subsection{Study Design}
\begin{figure*}[!htb]
     \centering
     \includegraphics[width=\textwidth]{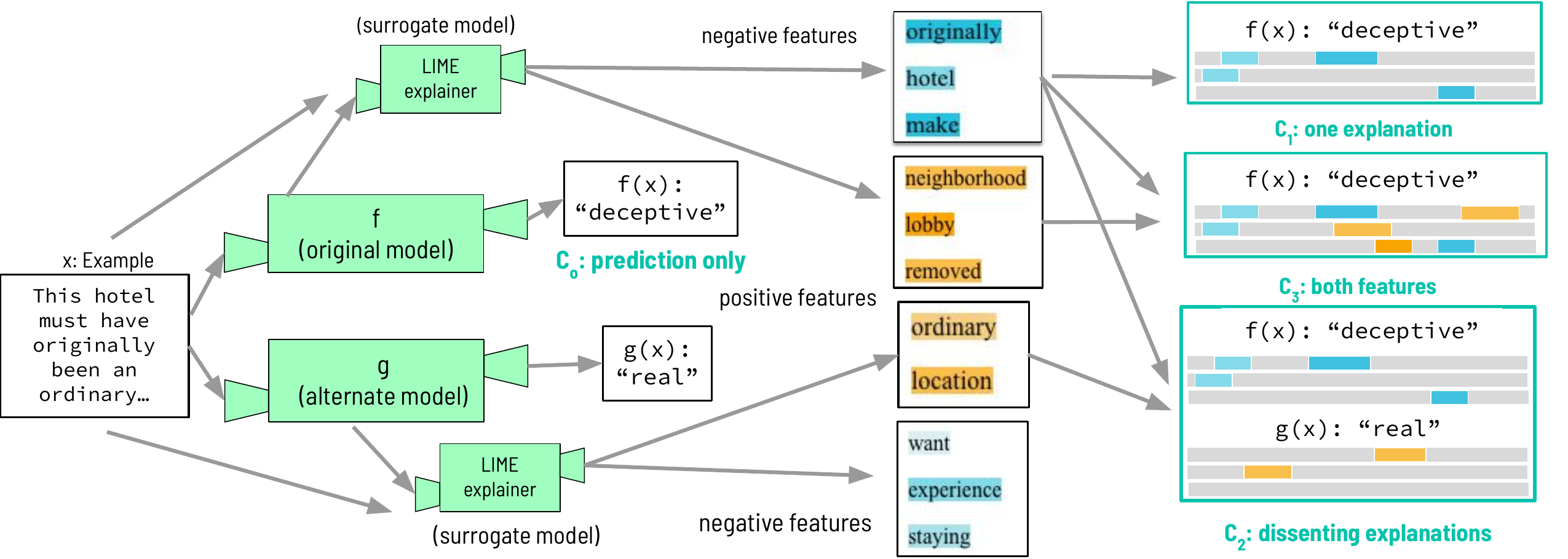}
     \caption{The process of generating explanations for the 4 conditions. The review text is for conceptual understanding only.}
     \label{fig:generatingconditions}
\end{figure*}

\paragraph{Task Selection} We focus on the setting of assistive AI: the setting where AI on average might perform better than humans but it is critical for humans to be the final decision maker. This is different from prior works, which focused on tasks either with ``verifiable'' answers given the explanation \cite{vasconcelos2022explanations} or tasks where humans and AI perform approximately equally in order to measure collaboration potential \cite{bansal2021does}. We use the term ``verifiable'' to describe tasks where explanations can verify that a model prediction is true (e.g. a maze task explanation provides a path to complete the maze or a math task explanation provides the step-by-step solution). We specifically consider explanations that are not verifiable proofs of the correct label but rather arguments for the model predictions, as these are the standard explanation forms available for complex model predictions \cite{burkart2020survey}. Since the effects of explanations differ widely from task to task \cite{wang2021explanations}, we focus on finding a specific, suitable task for studying dissenting explanations. We set the following requirements for our target task\footnote{We elaborate on these desiderata and discuss the candidate tasks we considered in the supplementary materials.}: 
\begin{enumerate}
    \item The human accuracy for the task must be less than the model accuracy.
    \item There must be room for model disagreement on the task; the AI model should not perform the task perfectly.
    \item There must be an objective correct label for the examples.
    \item AI explanations must be understandable without domain expertise and do not provide verifiable proof for the correct answer.
\end{enumerate}

\paragraph{Deceptive Reviews Task}
Based on our requirements, we decided to use the Chicago Deceptive Reviews dataset \cite{ott2011finding}: a dataset of 1600 one-paragraph reviews of hotels in Chicago, where half the reviews are genuine reviews from TripAdvisor, and the other half were written by crowd workers that have only seen the name and website of the hotel. The goal of the task is to distinguish between real and deceptive reviews; a prior study found that humans on their own get at most 62\% accuracy on this task, while a linear SVM achieved around 87\% accuracy \cite{lai2019human}. Furthermore, there exists a ground truth label: whether a review is deceptive or real. The explanations were in the form of highlighting the words selected by the feature attribution explainer; these words serve as an argument to the participant, convincing them to select a certain label without giving a complete proof of the correct answer. 

To test our hypotheses, we design a study in which human participants attempt to categorize these hotel reviews. Participants are presented with 20 hotel reviews, each of which is real or deceptive, and are instructed to decide which reviews are real. They are assisted by AI predictions or explanations, where the existence or type of explanation varies based on the condition participants are assigned to. Participants are warned in the beginning that the AI predictions are not always correct. Based on response quality in pilot studies, participants are also given a set of heuristics for identifying deceptive reviews, developed by prior work on this task \cite{lai2020chicago}. We also survey users, post-task, about task difficulty, the helpfulness of AI suggestions, and their trust in the AI suggestions. Finally, we included an optional open-ended question about how the AI suggestions helped them complete the task. 

\paragraph{Generating Explanations} To properly benchmark against prior work \cite{lai2019human}, we use the same linear SVM trained on TF-IDF of unigrams with English stop words removed as our reference model $f$ with 87\% accuracy. We also train an alternative 3-layer neural network model based on the exact same pre-processing which achieves 79\% accuracy as the alternative model $g$. We use LIME \cite{ribeiro2016should} to generate local explanations for each model using the Top-k features (k=15 was chosen based on the number of unique tokens per example for this dataset).

We used LIME for generating explanations in order to (1) remain consistent against prior work XAI studies in the text domain \cite{bansal2021does}, and (2) benchmark against \citep{lai2019human} who use linear model features (i.e. we compared our LIME features to the weights of the linear SVM model and found meaningful overlap\footnote{See details in in the supplementary materials}). To find dissenting explanations, we used examples in the test set where the neural network model disagreed with the linear SVM model. The two models disagreed on $10\%$ of examples (32 examples). We sub-sampled these examples for an even balance of examples that the linear SVM (the reference model) predicted correctly and incorrectly. This selection process yields a subset of examples that result in $\sim60\%$ accuracy; both for the reference model $f$ and human baseline.  

\paragraph{Conditions}

Each participant was presented with the same 20 reviews, along with the same 20 model predictions. The reference model $f$ predicted the incorrect label on 8 of the 20 reviews. We randomly assigned each participant to one of the following four conditions (Figure \ref{fig:generatingconditions}). Participants are not aware of the other possible conditions for the study.
\begin{itemize}
    \item $C_0$: Participants were presented with the AI prediction for each review, without any explanation.
    \item $C_1$: Participants were presented with the AI prediction for each review $f(x)$, along with a supporting explanation $e(x, f)_{f(x)}$. This means either the positive features were highlighted in orange, if the model predicted ``real," or the negative features were highlighted in blue, if the model predicted ``deceptive". 
    \item $C_2$: Participants were presented with both the explanation and the dissenting explanation. They received the same explanation as in $C_1$, followed by the line ``Another model predicts that this review is [real/deceptive]'' and the corresponding explanation for the dissenting model.
    \item $C_3$: Participants were presented with an explanation that more closely matched the original LIME output, which includes both positive and negative features. Each explanation started with the line ``The model predicts that this review is [real/deceptive]. It thinks the words in orange are evidence the review is real, while the words in blue are evidence it is deceptive." This was followed with the corresponding highlighted text.
\end{itemize}
We provided participants with training before the task began that was specifically tailored to the condition that each participant is assigned to. All other aspects of the survey, such as the format, the reviews, the predictions themselves, and the post-survey questions, were kept constant across all four conditions. To test \textbf{H1} and \textbf{H2}, we can compare $C_2$ to $C_1$. The baseline conditions we were also interested in understanding are provided by $C_0$ (prediction only) and $C_3$ (1 explanation containing both positive and negative features).
  
\paragraph{Participants}

Our study was run on Prolific and made available to all fluent English speakers that have at least a 95\% approval rate and have not answered any previous surveys (including pilot studies) we have posted. Participants were compensated $\$3.50$ USD for participating in the task and given an additional bonus of $\$1.00$ USD if they answered more than half the questions correctly. For the average completion time of $\sim 15$ minutes, this translates to a $\$18$ USD hourly rate. Three attention check questions were included in the study where participants were told explicitly to select a certain answer. We excluded answers from participants who failed more than one attention check but still compensated these participants according to IRB feedback. After excluding the failed attention checks, there were $N = 178$ submissions in our analysis, with approximately $45$ submissions per condition\footnote{Our deceptive reviews task obtained IRB exemption approval and our study was pre-registered}. Our sample size was calculated based on pilot studies. 

\subsection{Results}
\begin{figure*}[h]
     \centering
          \begin{subfigure}[b]{0.33\textwidth}
         \centering
         \includegraphics[width=\textwidth]{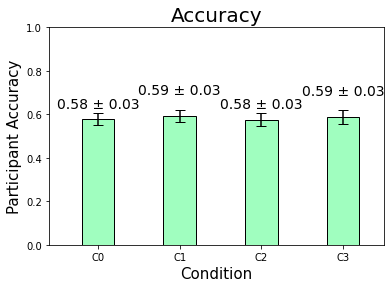}
         \caption{Accuracy}
         \label{fig:results-acc}
     \end{subfigure}
     \begin{subfigure}[b]{0.33\textwidth}
         \centering
         \includegraphics[width=\textwidth]{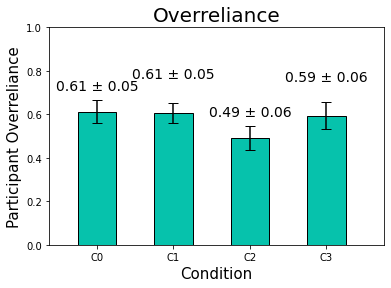}
         \caption{Overreliance}
         \label{fig:results-over}
     \end{subfigure}
     \begin{subfigure}[b]{0.33\textwidth}
         \centering
         \includegraphics[width=\textwidth]{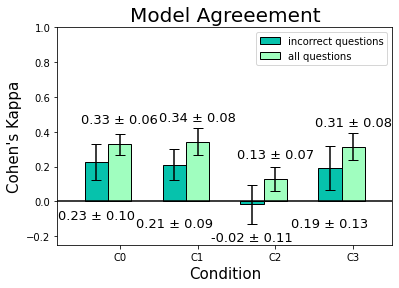}
         \caption{Agreement}
         \label{fig:results-agreement}
     \end{subfigure}
    \caption{(a) Accuracy, (b) Overreliance, and (c) Agreement (Cohen's $\kappa$ score) for each experimental condition demonstrating that dissenting explanations $C_2$ significantly reduce overreliance without reducing overall accuracy. Error bars represent $95\%$ confidence intervals of the mean across participants (N=178).}
    \label{fig:results}
    \end{figure*} 

For each participant in the study, we measured their \textbf{accuracy} as the fraction of reviews they categorized correctly, out of the 20 total reviews. We measured \textbf{overreliance} as the fraction of reviews they agreed with the model prediction on, out of the 8 reviews the model predicted incorrectly. These results, averaged over users of each of the four conditions, are displayed in Figure \ref{fig:results-acc} and Figure \ref{fig:results-over}. Since our task involves binary labels, we account for random agreement by also measuring Cohen's $\kappa$ between a participant and the model's predictions in Figure \ref{fig:results-agreement} \cite{mchugh2012interrater}.  

Using a one-way ANOVA test, we find that accuracy does not differ significantly, but overreliance ($p=0.007$) and Cohen's $\kappa$ ($p<0.001$) scores do differ across conditions. We then perform one-tailed $t$-tests between conditions to test our specific hypotheses. To analyze H1, we examined 8 questions that the model predicted incorrectly; both Cohen's $\kappa$ and the overreliance were not greater in $C_0$ than $C_1$.

We find that our results support our main hypothesis (H2): providing participants with both a supporting and dissenting explanation ($C_2$) significantly reduces overreliance as compared to just a single explanation ($C_1$) ($p<0.001$, Figure \ref{fig:results-over}). We also observe that for human-model agreement, as measured by Cohen's $\kappa$, dissenting explanations in condition $C_2$ also give a significantly lower agreement with model predictions than just a single explanation $C_1$ ($p<0.001$, Figure \ref{fig:results-agreement}). However, since the dissenting explanation condition $C_2$ does \textit{not} significantly reduce accuracy ($p>0.05$, Figure \ref{fig:results-acc}), this suggests that dissenting explanations reduce human-model agreement as well as overreliance. If the decrease in agreement only resulted in underreliance (i.e. human disagreement with the model prediction when the model is correct, also increased due to decreased agreement), we would observe a drop in overall accuracy.


To understand whether the reduction in overreliance arises from simply the presentation of more information or increased cognitive load, we can compare $C_2$ with $C_3$, where participants saw both the positive and negative features from a single model explanation. We find that condition $C_3$ produced significantly higher overreliance compared to the $C_2$ condition ($p=0.009$). This shows that dissenting explanations may help calibrate trust; there is a significant difference in how humans react to positive evidence from one model and negative evidence from another, as opposed to positive and negative evidence from a singular model in this deception labeling task.

\begin{figure}
     \centering
     \includegraphics[width=0.4\textwidth]{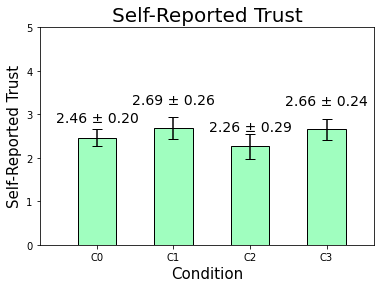}
     \caption{Level of trust reported by participants on a scale of (1) ``\textit{not at all}'' to (5) ``\textit{a great deal}''. The level of trust was significantly lower for $C_2$ (dissenting explanations condition)}
     \label{fig:trust}
 \end{figure}

\paragraph{Qualitative Analysis}
Participants were asked to report their trust in the AI predictions, on a 5-point scale from ``\textit{not at all}'' to ``\textit{a great deal}''. The reported trust matched the trend of the overreliance scores across the 4 conditions, where the average reported trust in the model predictions was lowest in the dissenting explanations condition, and higher in the other 3 conditions (Figure \ref{fig:trust}). This was reflected in participant comments; one comment for $C_0$ was ``\textit{If i was on the fence on wheter it was fake or not i tried to listen to the AI suggestion}", and many others had a similar sentiment. For condition $C_2$, there were many comments saying they distrusted the AI suggestions, and a few saying that they followed the suggestion with the more-highlighted paragraph. Similarly, in $C_3$, there were many comments such as ``\textit{It helped me to easily identify the ammount of key words of each type.}''  Thus, the participants' beliefs about the study generally reflected the quantitative results we found for each of the explanation conditions.
\section{Finding Dissenting Explanations}
Motivated by the utility of dissenting explanations, we present methods for producing disagreement in models. While many techniques may be developed to create model multiplicity, we focus on simple objective-based approaches that are effective for our task of interest similar to techniques from robustness and diverse ensembles \cite{pang2019improving, rame2021dice}\footnote{Additional experiments for these techniques on other datasets are included in the supplementary materials}. While prior works have focused on predictive multiplicity, a clear mapping between predictive and explanation multiplicity has not been presented. In this section, we present and compare methods for increasing predictive multiplicity through the lens of explanations. 
\subsection{Global Disagreement: A Model Agnostic Approach} 
We consider the setting where we have access to a reference model $f$ and the training set. Our goal is to suggest simple but intuitive methods to train a model $g$ which will disagree with $f$ as much as possible on a subsequent test set.  
\begin{problem}{\textbf{Global Disagreement}}
Given reference model $f$ and training data $D$, find some $g$ such that $\delta_D(f, g)$ (Definition \ref{def:global_disagree}) is maximized while  $\Err_{D_{\text{test}}}(f) \approx \Err_{D_{\text{test}}}(g)$. 
\label{prob:global}
\end{problem}
\paragraph{Regularization (\textsc{Reg})}  First, we consider a regularization approach to penalize similarities between a \textit{fixed} reference model $f$ predictions and the current model $g$. Specifically, one empirical loss we can minimize is: 
\begin{equation}
    L(x, y, f) = \frac{1}{n}\sum_{i=1}^{n}l(g(x_i), y_i) + \frac{\lambda}{n}\sum_{i=1}^{n}\mathds{1}[f(x_i) \ne g(x_i)]
\end{equation}
However, since the indicator function is not continuous and non-differentiable, we modify the objective to be: 
\begin{equation}
    L(x, y, f) = \frac{1}{n}\sum_{i=1}^{n}l(g(x_i), y_i) + \frac{\lambda}{n}\sum_{i=1}^{n}l(g(x_i), \overline{f(x_i)})
\end{equation}
We consider the binary classification setting and set $l$ to be the binary cross entropy loss and use the inverse predictions of $f$ to maximize disagreement between $f$ and $g$.  

\paragraph{Reweighting (\textsc{Weights)}}
Leveraging intuition from boosting, another approach to learning a maximally differing classifier is to upweight examples that our reference predictor gets wrong. Our approach differs from traditional boosting in that we are comparing explanations between resulting models instead of combining model outputs for a single prediction. Formally, the reweighting objective is as follows: 
\begin{align}
    L(x, y, f) &= \frac{1}{n}\sum_{i=1}^{n}w_il(g(x_i), y_i) \\
    w_i &= 1 + \lambda \mathds{1}[f(x_i) \ne y_i)]
\end{align}
\begin{remark}
When $l(x, y) = \mathds{1}[x \ne y]$, in the binary setting this reweighting objective is equivalent to the above regularization objective. 
\end{remark}
\begin{table}[h]
    \begin{subtable}[h]{0.45\textwidth}
        \centering
        \begin{tabular}{|l | l | l| l|}
        \hline
        $\lambda$ & Accuracy & Disagreement & Corr.\\
        \hline 
        0.0 & 0.889 $\pm$ .010 & 8.66 $\pm$ 0.6 \% & 40.1 \%\\
        0.1 & 0.883 $\pm$ .017 & 8.75 $\pm$ 0.5 \% & 38.9 \%\\
        0.25& 0.859 $\pm$ .021 & 10.9 $\pm$ 3.4 \% & 34.2 \% \\ 
        \textbf{0.5} & \textbf{0.807 $\pm$ .017} & \textbf{16.6 $\pm$ 2.3} \% & \textbf{35.7 \%} \\
        \hline
       \end{tabular}
       \caption{\textsc{Reg} objective (batch size 10)} 
       \label{tab:reg}
    \end{subtable}
    \hfill
    \begin{subtable}[h]{0.45\textwidth}
        \centering
        \begin{tabular}{|l | l | l |l|}
        \hline
        $\lambda$ & Accuracy & Disagreement & Corr. \\
        \hline 
        0  & 0.859 $\pm$ .019 & 8.68 $\pm$ 0.7 \% & 28.4\%\\
        1  & 0.865 $\pm$ .014 & 8.56 $\pm$ 1.2 \% & 30.5\%\\
        10 & 0.854 $\pm$ .008 & 10.8 $\pm$ 1.5 \% & 35.3\%\\
        \textbf{50} & \textbf{0.826 $\pm$ .018 }& \textbf{14.9 $\pm$ 0.7} \% & \textbf{40.1\%}\\
        \hline
        \end{tabular}
        \caption{\textsc{Weights} objective (batch size 100)}
        \label{tab:weights}
     \end{subtable}
     \caption{Comparison of proposed methods to elicit predictive multiplicity against a $88\%$ accuracy reference model ($f$). \textsc{Weights} requires a larger batch size to ensure an incorrect example is included, Corr. indicates the percentage of $f$'s incorrect predictions which were corrected by $g$.}
\end{table}
\paragraph{Experiment Results} First, we compare predictive multiplicity induced by both methods on the deceptive reviews dataset and use the same reference model $f$, a linear-SVM, from our study. For all experiments in this section, we train a neural network $g$ with a single hidden layer with the same features as the reference model $f$. The results presented are averaged over 5 different random seeds. Table \ref{tab:reg} summarizes the overall model accuracy, the percentage of examples $f$ and $g$ disagreed on, and the percentage of examples that were incorrectly predicted by $f$ but rectified by $g$ computed over a held-out test set. As $\lambda$ increases, the number of conflicting prediction examples also increases\footnote{Training with the \textsc{Reg} objective using larger $\lambda$ (e.g. $\lambda \ge 1$) resulted in instabilities for a variety of hyperparameters.}. However, this effect might be due to $g$ simply getting more examples wrong. Thus, it is important to measure the number of $f$'s incorrect predictions that are corrected by $g$. Of the total 38 examples in the test set that $f$ predicts incorrectly, the percentage of corrected examples reduces slightly as disagreement increases. Table \ref{tab:weights} summarizes the effectiveness of using the \textsc{Weights} objective in creating model predictive disagreement. Disagreement is achieved without as much sacrifice in overall accuracy and both the percentage disagreement and corrected samples are high at larger $\lambda$ values. 

\paragraph{Explanation Disagreement}

\begin{figure}[!h]
     \begin{subfigure}[b]{0.42\textwidth}
         \centering
         \includegraphics[width=\textwidth]{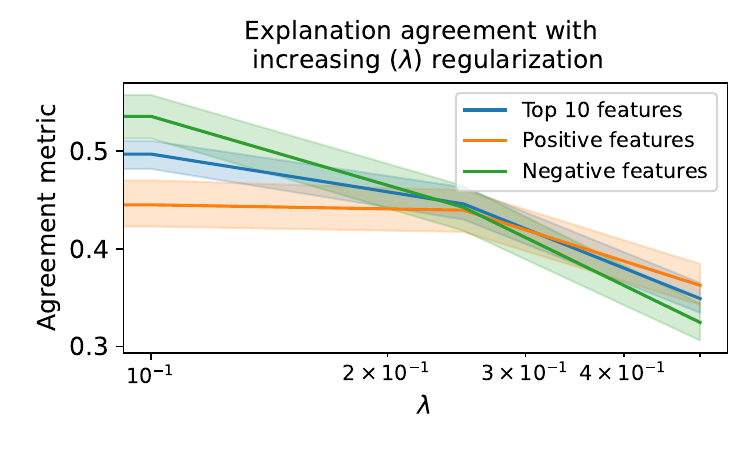}
         \caption{Explanation Agreement from \textsc{Reg}}
         \label{fig:reg_exp_agree}
     \end{subfigure}
    \hfill
     \begin{subfigure}[b]{0.42\textwidth}
         \centering
         \includegraphics[width=\textwidth]{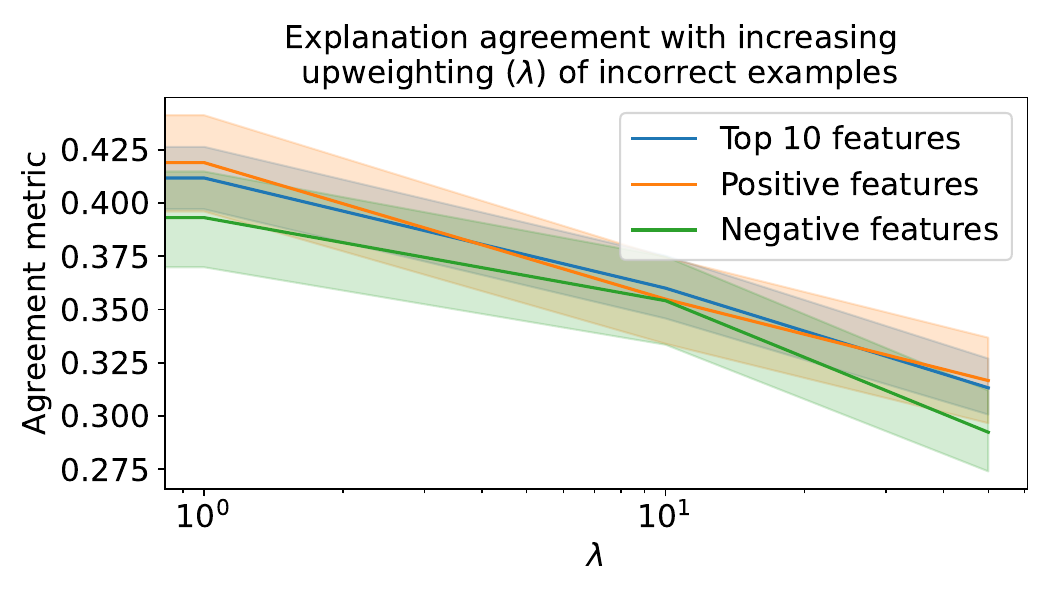}
         \caption{Explanation agreement from \textsc{Weights}}
         \label{fig:reweigh_exp_agree}
     \end{subfigure}
    \caption{As we emphasize the importance of model predictive disagreement through increasing $\lambda$, the agreement between explanations as measured by the overlap in top features also decreases.}
    \label{fig:exp_agree}
\end{figure}

For comparing dissenting explanations to the original explanation, we use a similar set of metrics as explanation agreement \cite{krishna2022disagreement}. 
We consider three agreement metrics \textsc{TopK},  \textsc{TopKPos}, and  \textsc{TopKNeg}: 
\begin{align}
     \textsc{TopK} &= \frac{|\topt_k(e(x, f)) \cap \topt_k(e(x, g))|}{|\topt_k(e(x, f)) \cup \topt_k(e(x, g))|} \\
 \textsc{TopKPos} &= \frac{|\topt_k(e(x, f))_+ \cap \topt_k(e(x, g))_+|}{|\topt_k(e(x, f))_+ \cup \topt_k(e(x, g))_+|}
\end{align}
\textsc{TopKNeg} is just \textsc{TopKPos} with negative prediction features instead of positive. To measure explanation agreement, we evaluate models at different $\lambda$ for both \textsc{Reg} (Figure \ref{fig:reg_exp_agree}) and \textsc{Weights} (Figure \ref{fig:reweigh_exp_agree}). For all three metrics, as $\lambda$ increases, the explanation agreement also reduces. Although these results are unsurprising, they are compelling in illustrating that creating predictive multiplicity also in turn produces explanation multiplicity. Moreover, since a good portion of examples that the reference model classified incorrectly were rectified in our alternative models, this explanation multiplicity allows dissenting explanations to aid human judgment and reduce overreliance.  

\subsection{Local Disagreement: Generating a Dissenting Explanation for Any Input}
\begin{table}[h]
    \begin{subtable}[h]{0.45\textwidth}
        \centering
        \begin{tabular}{|l | l | l| l|}
        \hline
        $|D|$ & Success Rate & \textsc{TopK} Agree. & Acc.\\
        \hline 
        1280 & 0.543 $\pm$ .249 & 0.756 $\pm$ .131 & 0.880\\
        640 & 0.723 $\pm$ .200 & 0.464 $\pm$ .122 & 0.889\\
        320 & 0.910 $\pm$ .082 & 0.352 $\pm$ .111 & 0.844\\ 
        160 & 0.987 $\pm$ .013 & 0.275 $\pm$ .115 & 0.780\\
        80  & 1.000 $\pm$ .000 & 0.227 $\pm$ .103 & 0.675\\
        \hline
       \end{tabular}
       \caption{SVM} 
       \label{tab:svm-local}
    \end{subtable}
    \hfill
    \begin{subtable}[h]{0.45\textwidth}
        \centering
        \begin{tabular}{|l | l | l |l|}
        \hline
        Iter. & Freq. & \textsc{TopK} Agree. & Acc. \\
        \hline 
        $<5$    & 19.7\% & 0.946 $\pm$ .091 & 0.902 \\
        5-10  & 20.9\%  & 0.878 $\pm$ .113 & 0.892\\
        10-15 & 18.1\% & 0.786 $\pm$ .117 & 0.886 \\
        15-20 & 19.1\% & 0.770 $\pm$ .159 & 0.883\\
        $>20$   & 22.2\% & 0.782 $\pm$ .114 & 0.869\\
        \hline
        \end{tabular}
        \caption{Neural Network}
        \label{tab:nn-local}
     \end{subtable}
     \caption{(a) Success rate, \textsc{TopK} agreement ($f$ vs $g$), and test set accuracy of $g$ when adding a test instance to the training set for a flipped prediction. As dataset size decreases, instances are more likely to be successfully predicted as the opposite class. (b) Distribution of training iterations required and resulting \textsc{TopK} agreement ($f$ vs $g$) and accuracy of $g$, for a neural network model that is retrained on a single test instance. Errors reported are standard deviation for all values and variance for success rate (Bernoulli). All test set accuracy errors are $<0.02$.}
\end{table}
While the techniques we presented increase model disagreement on the test data only with the training data, the total coverage only spans $<30\%$ of points of the test set. We now consider an alternative problem formulation where the test instance for which we want to achieve a different prediction is given. This allows us to produce a dissenting explanation for any input example in the reference model.
\begin{problem}{\textbf{Local Disagreement}}
Given reference model $f$, training data $D$, and a test instance $x$,  find some $g$ where ${f(x) \ne g(x)}$ where  $\Err_{D_{\textup{test}}}(f) \approx \Err_{D_{\textup{test}}}(g)$. 
\label{prob:local}
\end{problem}
Here we know the exact test instance for which we want a different prediction. We both consider the scenario of additional access to only the training set and the scenario of additional access to only the trained model. For the former, we use a linear SVM to demonstrate the efficacy of flipping the test instance label by reducing the training size $|D|$. As the training data size decreases, the influence of the test instance increases, improving the success rate of generating a different prediction for that instance. Table \ref{tab:svm-local} shows that Problem \ref{prob:local} can be solved with high probability (i.e. $>90\%$) without sacrificing significant model accuracy for the deceptive reviews task. 
For the scenario where we only have additional access to the trained model (i.e. a neural network fitted to the training set), we retrain directly to minimize the loss on the test instance and measure how many iterations are required to change the label. Table \ref{tab:nn-local} describes the distribution over iterations required to flip an example label. This method is also effective in flipping the label for roughly $80\%$ of the test set examples while still maintaining $\sim 88\%$ accuracy.

\section{Discussion}
In this work, we take a holistic approach by first motivating the need for dissenting explanations through a human study to measure overreliance. For our deceptive reviews task, a task with a ground truth label but no method for direct verification, we demonstrate the utility of dissenting explanations in reducing overreliance. Our results complement existing work on the benefits of explanations \cite{vasconcelos2022explanations} by exploring more ambiguous tasks. 

After finding that overreliance can be reduced by dissenting explanations which argue against a model prediction and explanation, we defined types of disagreement and presented simple but effective heuristics for eliciting such disagreements. We show that generating disagreement in predictions is sufficient for generating different explanations. Our work serves as a first step in connecting human interaction and computational challenges in treating explanations as arguments for AI-assisted decision-making. 

\paragraph{Limitations and Future Work} Our work, on a single task, cannot make claims about the effectiveness of dissenting explanations in general. The subset of examples we examine is difficult for the reference model, thus providing a single explanation does not significantly assist human decision-making; future work should use a stronger reference model with more examples. A promising direction of future work is to explore what tasks dissenting explanations best aid and other types of dissenting explanations involving counterfactual explanations. 

\section*{Acknowledgments}
This research was supported by the Simons collaboration on the theory of algorithmic fairness and the Simons Foundation Investigators Award 689988. Also thanks to Lindsay Popowski and Aspen Hopkins for the insightful discussions on experimental design.

\bibliography{main}

\newpage
\appendix
\section{Additional Experiments}
In our main text, we focus on the Hotel reviews dataset we use for our study. To demonstrate the effectiveness of our proposed approaches, we run the same set of experiments both for global and local disagreement on the UCI Adult dataset from Folktables \cite{ding2021retiring}. We use the Income task for Hawaii from 2018. 

\begin{table}[h]
    \begin{subtable}[h]{0.45\textwidth}
        \centering
        \begin{tabular}{|l | l | l| l|}
        \hline
        $\lambda$ & Accuracy & Disagreement & Corr.\\
        \hline 
        0.0  & 0.754 $\pm$ .002 & 6.12 $\pm$ 0.6 \% & 13.5 \%\\
        0.10 & 0.756 $\pm$ .006 & 7.42 $\pm$ 0.8 \% & 16.5 \%\\
        0.25 & 0.753 $\pm$ .011 & 10.8 $\pm$ 0.9 \% & 22.7 \% \\ 
        0.50 & 0.706 $\pm$ .004 & 27.3 $\pm$ 2.6 \% & 43.6 \% \\
        \hline
       \end{tabular}
       \caption{\textsc{Reg} objective (batch size 10)} 
       \label{tab:reg_income}
    \end{subtable}
    \hfill
    \begin{subtable}[h]{0.45\textwidth}
        \centering
        \begin{tabular}{|l | l | l |l|}
        \hline
        $\lambda$ & Accuracy & Disagreement & Corr. \\
        \hline 
        0   & 0.766 $\pm$ .002 & 9.10 $\pm$ 0.5 \% & 22.1\%\\
        0.5 & 0.760 $\pm$ .006 & 13.1 $\pm$ 1.2 \% & 30.2\%\\
        1.0 & 0.748 $\pm$ .007 & 18.0 $\pm$ 1.7 \% & 37.6\%\\
        5.0 & 0.560 $\pm$ .002 & 56.3 $\pm$ 6.8 \% & 70.0\%\\
        \hline
        \end{tabular}
        \caption{\textsc{Weights} objective (batch size 100)}
        \label{tab:weights_income}
     \end{subtable}
     \caption{Comparison of proposed methods to elicit predictive multiplicity against a $88\%$ accuracy reference model ($f$) on the Adult Income Dataset. \textsc{Weights} requires a larger batch size since the reference model is very accurate. Corr. indicated the percentage of $f$'s incorrect predictions which were corrected by $g$.}
     \label{tab:temps}
\end{table}
\subsection{Global Disagreement}
\paragraph{Experiment Results} Similar to the reviews dataset, we see that both methods are fairly effective for inducing model disagreement without sacrificing too much model accuracy. These experiments were also for a linear SVM ref model $f$ against a neural network model. 

\paragraph{Explanation disagreement}

\begin{figure}[h]
     \begin{subfigure}[b]{0.47\textwidth}
         \centering
         \includegraphics[width=\textwidth]{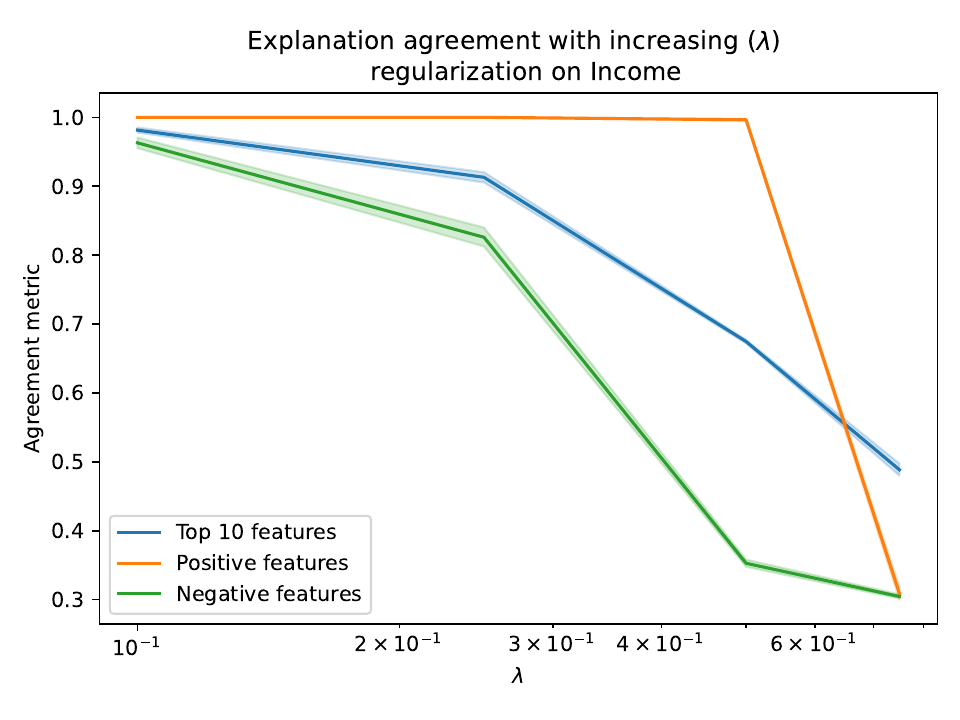}
         \caption{Explanation Agreement from \textsc{Reg}}
         \label{fig:reg_exp_agree_income}
     \end{subfigure}
    \hfill
     \begin{subfigure}[b]{0.47\textwidth}
         \centering
         \includegraphics[width=\textwidth]{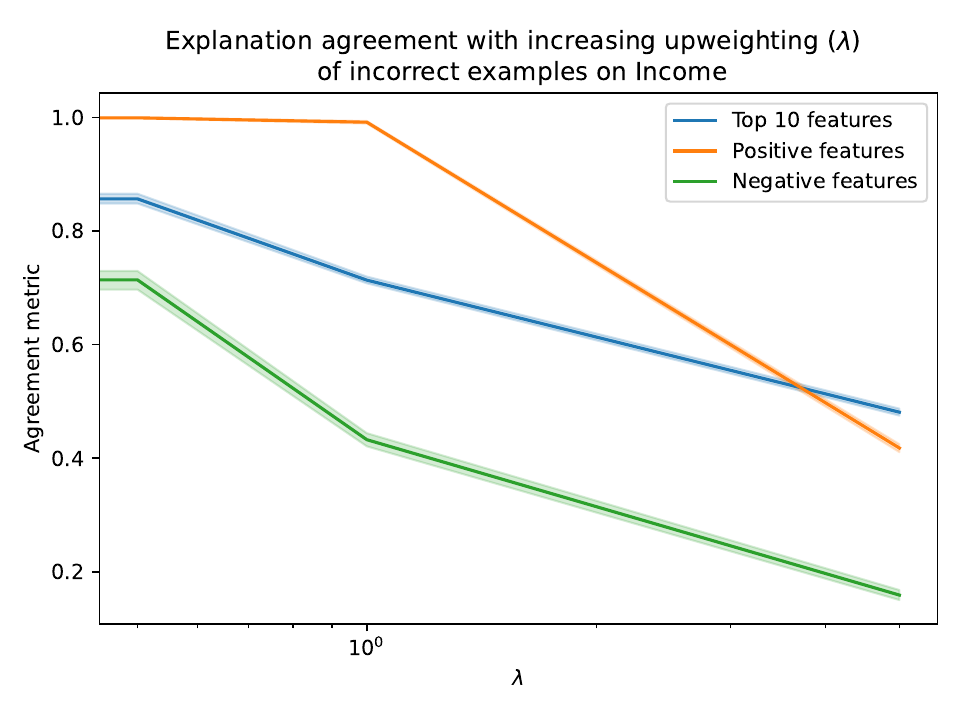}
         \caption{Explanation agreement from \textsc{Weights}}
         \label{fig:reweigh_exp_agree_income}
     \end{subfigure}
    \caption{As we emphasize the importance of model predictive disagreement through increasing $\lambda$, the agreement between explanations as measured by the overlap in top features also decreases.}
    \label{fig:exp_agree_income}
\end{figure}
Similarly, the amount of explanation disagreement also increases with increasing $\lambda$ for both methods. 

\subsection{Local model disagreement: generating a dissenting explanation for any input}
\begin{table}[h]
    \begin{subtable}[h]{0.45\textwidth}
        \centering
        \begin{tabular}{|l | l | l| l|}
        \hline
        $|D|$ & Success Rate & \textsc{TopK} Agree. & Acc.\\
        \hline 
        6185 & 0.987 $\pm$ .013 & 0.749 $\pm$ .114 & 0.771\\
        3201 & 0.992 $\pm$ .008 & 0.776 $\pm$ .157 & 0.767\\
        1601 & 0.995 $\pm$ .005 & 0.825 $\pm$ .167 & 0.760\\ 
        801 & 0.996 $\pm$ .004 & 0.978 $\pm$ .008 & 0.756\\
        401 & 0.999 $\pm$ .001 & 0.709 $\pm$ .160 & 0.734\\
        \hline
       \end{tabular}
       \caption{SVM} 
       \label{tab:svm-local-adult}
    \end{subtable}
    \hfill
    \begin{subtable}[h]{0.45\textwidth}
        \centering
        \begin{tabular}{|l | l | l |l|}
        \hline
        Iter. & Freq. & \textsc{TopK} Agree. & Acc. \\
        \hline 
        $<5$    & 24.8\% & 0.979 $\pm$ .081 & 0.763 \\
        5-10  & 22.6\% & 0.970 $\pm$ .095 & 0.753 \\
        10-15 & 14.9\% & 0.973 $\pm$ .090 & 0.741 \\
        15-20 & 9.8\% & 0.982 $\pm$ .075 & 0.726 \\
        $>20$   & 21.7\% & 0.61 $\pm$ .106 & 0.713 \\
        \hline
        \end{tabular}
        \caption{Neural Network}
        \label{tab:nn-local-adult}
     \end{subtable}
     \caption{(a) Success rate, \textsc{TopK} agreement between $f$ and $g$, and test set accuracy of $g$ when adding a test instance to the training set for a flipped prediction for the Adule Income dataset. As dataset size decreases, the test instance is more likely to be successfully predicted as the opposite class. (b) Training iterations required to find $g$, \textsc{TopK} agreement between $f$ and $g$, and test set accuracy of $g$, a neural network model that is retrained on the test instance. With more training iterations on the single instance, the prediction for the instance is more likely to flip. Errors reported are standard deviation for all values and variance for Success rate (Bernoulli).}
     \label{tab:temps_adult}
\end{table}
Similar to the experiments in the main text, we see can find a learning rate where retraining on a single example leads to success in flipping the label of a test example without sacrificing too much accuracy in the neural network.

\section{Human Experiment Details}

\subsection{Task Selection} In order to properly analyze how dissenting-explanations affect human overreliance, the task we use must allow humans to learn from AI explanations, without explanations completely solving the problem for the human participants. Prior works focused on tasks either with verifiable answers given the explanation \cite{vasconcelos2022explanations} or tasks where humans and AI perform approximately equally in order to measure collaboration potential \cite{bansal2021does}. We focus on assistive AI: the setting where AI on average might perform better than humans but it is critical for humans to be the final decision maker. Furthermore, we specifically consider explanations that cannot fully verify the model prediction but rather makes an argument for the model's prediction. 
Thus, we had the following criteria in selecting a task: \begin{enumerate}
    \item \textit{The human accuracy for the task must be less than the model accuracy.} Assistive AI should provide explanations as an opportunity for humans to learn from the model decision and improve their choices; humans should perform worse on the task if they entirely ignore the model prediction and explanation.
    \item \textit{There must be room for model disagreement on the task; the AI model should not perform the task perfectly.} To test overreliance, the task must be one where completely relying on AI predictions is not an optimal strategy. Moreover, there must be enough examples where two models disagree in order to provide dissenting-explanations for the study.
    \item \textit{There must be an objective correct label for the examples.} \\ Since the purpose of the study is to measure human trust in different types of AI explanations, a more subjective classification task might cause results to be complicated by trust or distrust in a more subjective true label. Furthermore, in tasks like income prediction and loan default, two individuals with the same features may have different outcomes. It is difficult the correctness of (somewhat) randomized outcomes. 
    \item \textit{AI explanations must be understandable for the participants, without providing complete proof of the correct answer.} We examine settings where there is an objective ground truth but explanations argue for rather than directly verify a prediction (e.g. explanations for tasks like math questions and mazes verify rather than argue for a prediction). Furthermore, understandable also implies that domain expertise is not necessary for decision-making.  
\end{enumerate}

With these criteria in mind, we review existing tasks where the value of explanations has been tested. Table \ref{tab:my_label} summarizes the explainability tasks that have been studied within the context of AI decision-making as well as the reasons that these tasks did not match our criteria. The only prior work presenting a task that matched our criteria is by \citet{buccinca2020proxy}. They provide other examples in as inductive support for an AI's decision as well as ingredient lists for deductive support. For the purpose of our study, we focus on deductive support from explanations with are generated by real-world explainability tools. It is unclear what model disagreement would look like on this task so we leave the study of dissenting explanations on this task to future work. 

Thus we end up with only deceptive reviews as the task with precedent in prior work which fits all of our criteria. Moreover, the human-only baseline on this task has already been tested to be 51.1\% \cite{lai2019human}. 

\begin{table*}[t]
    \centering
    \begin{tabular}{|l|p{3.5in}|}
    \hline
        Task & Reason for Elimination\\ 
    \hline
        Amazon Reviews \citep{bansal2021does} & Machine accuracy worse than humans (Reason 1) \\ 
        LSAT \citep{bansal2021does} & Machine accuracy worse than humans (Reason 1)\\ 
        Income Prediction \citep{zhang2020effect} & No objectively correct label (Reason 3) \\ 
        Loan Defaults \citep{green2019principles} & No objectively correct label (Reason 3) \\ 
        Hypoxemia risk \citep{lundberg2018explainable} & Explanation understanding requires domain expertise (Reason 4) \\ 
        Maze \citep{vasconcelos2022explanations} & No significant model disagreement and explanation provides proof/verification for prediction (Reason 2 \& 4)\\
        Deceptive Reviews \citep{lai2019human} & N/A (\textit{Chosen Task})\\ \hline 
    \end{tabular}
    \caption{Classification tasks from prior work on explanations we considered and reasons we eliminated the task as a possible prototype for dissenting explanations}
    \label{tab:my_label}
\end{table*}
\subsection{Participant Details}
A pilot study was first done with 6 participants in each condition; this gave us feedback on the task, and caused us to slightly modify the explanations (by improving the highlighting and removing stop words from the explanation) for the final task. 

After this, a second pilot study was done with 20 participants in each condition, and a power analysis was done on this pilot study, which found that 40 participants per condition $(N = 160)$ was sufficient for getting significant results for our hypothesis. We send the survey out to $N=200$ participants to account for failed attention checks and other participation issues. As per IRB request, we still compensated participants if they failed attention checks.

\subsection{Survey Details}
\begin{figure}[h]
\centering
    \begin{subfigure}[h]{0.5\textwidth}
        \centering
        \includegraphics[width=\textwidth]{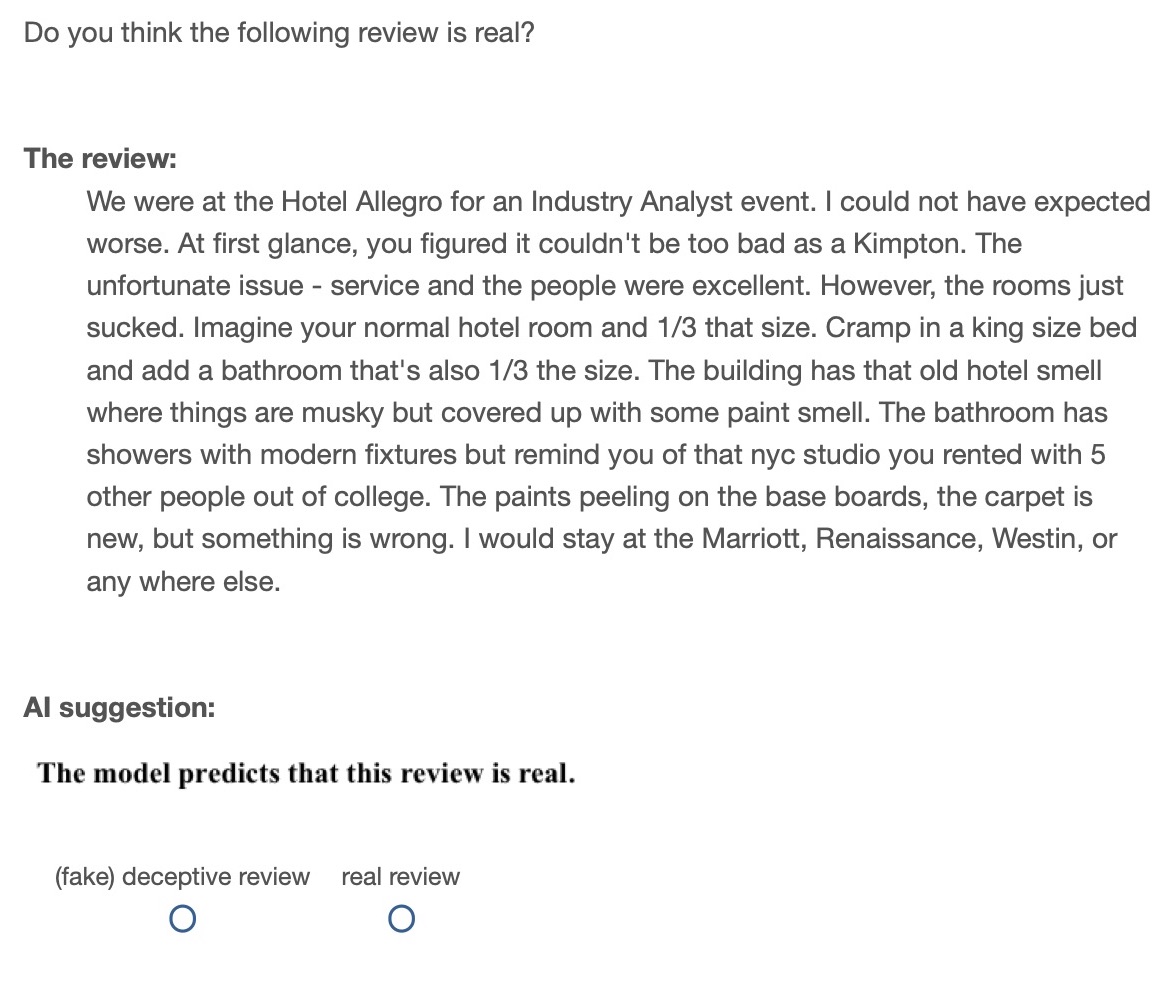}
        \caption{An example question presented to participants in condition $C_0$}
        \label{fig:q0}
    \end{subfigure}
    \hfill
    \begin{subfigure}[h]{0.5\textwidth}
        \centering
        \includegraphics[width=\textwidth]{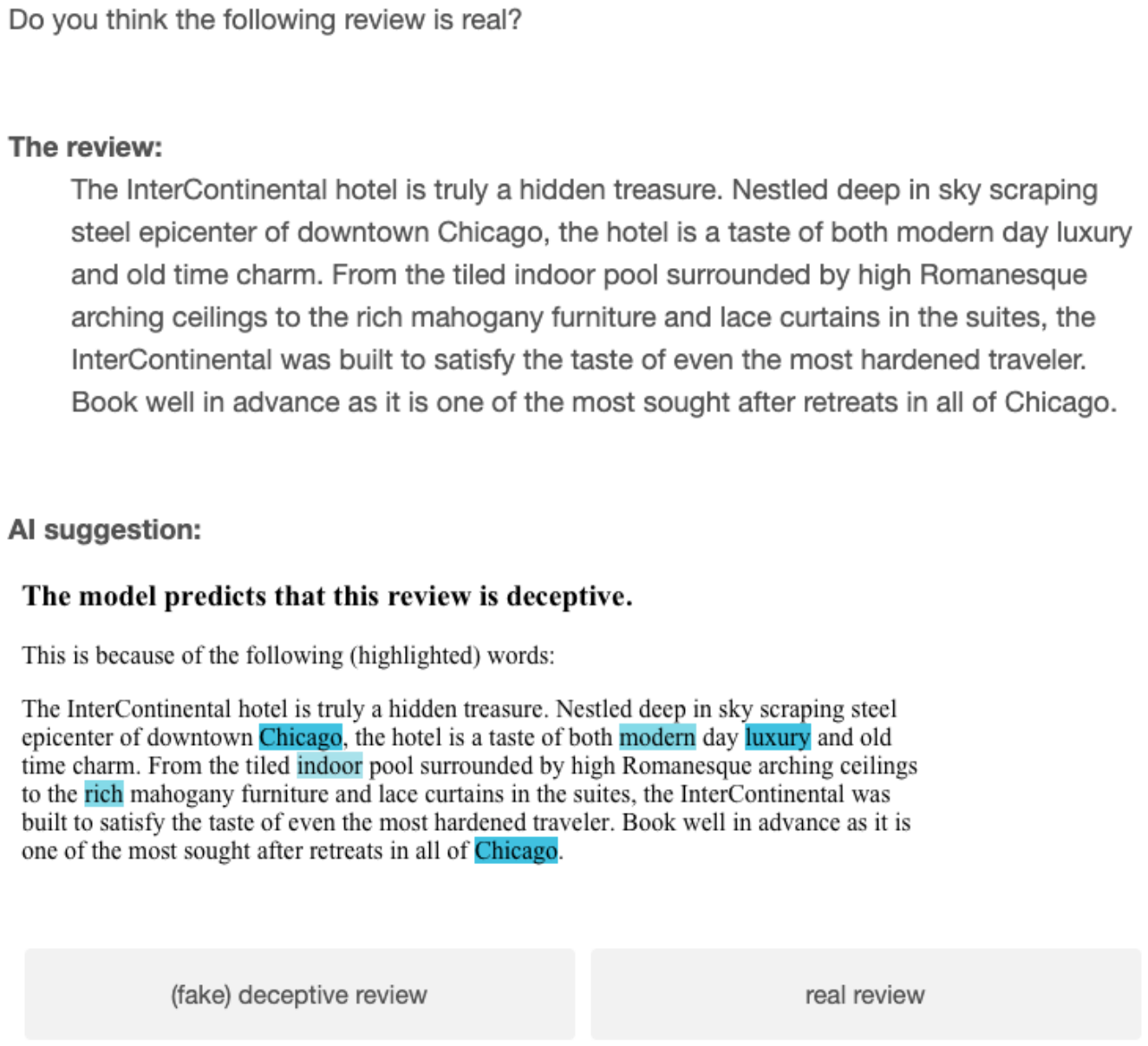}
        \caption{an example question presented to participants in condition $C_1$}
        \label{fig:q1}
    \end{subfigure}
    \caption{An example question from the survey for conditions $C_0$ and $C_1$}
\end{figure}
\begin{figure}[h]
    \begin{subfigure}[h]{0.5\textwidth}
        \centering
        \includegraphics[width=\textwidth]{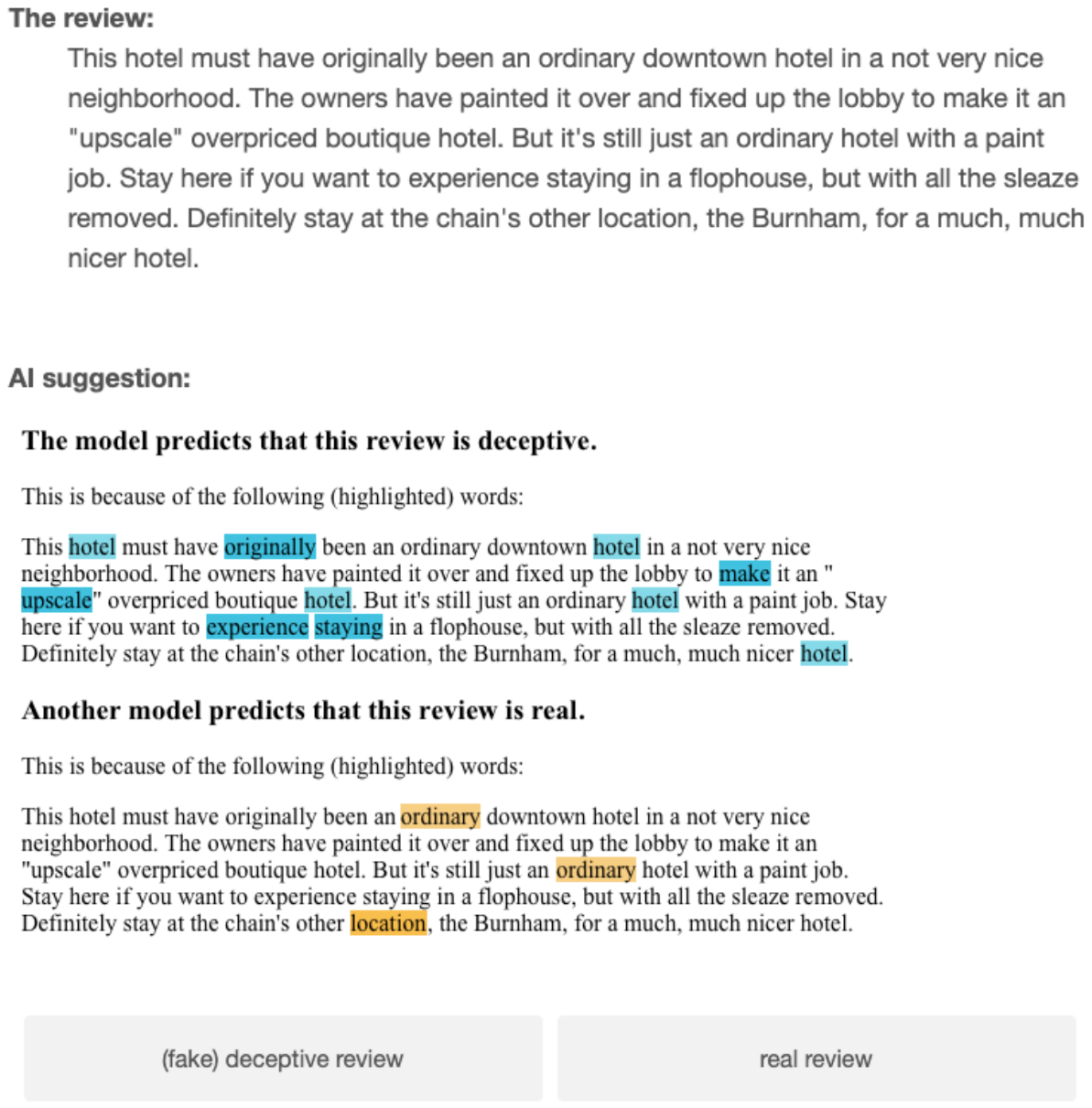}
        \caption{an example question presented to participants in condition $C_2$}
        \label{fig:q2}
    \end{subfigure}
    \begin{subfigure}[h]{0.5\textwidth}
        \centering
        \includegraphics[width=\textwidth]{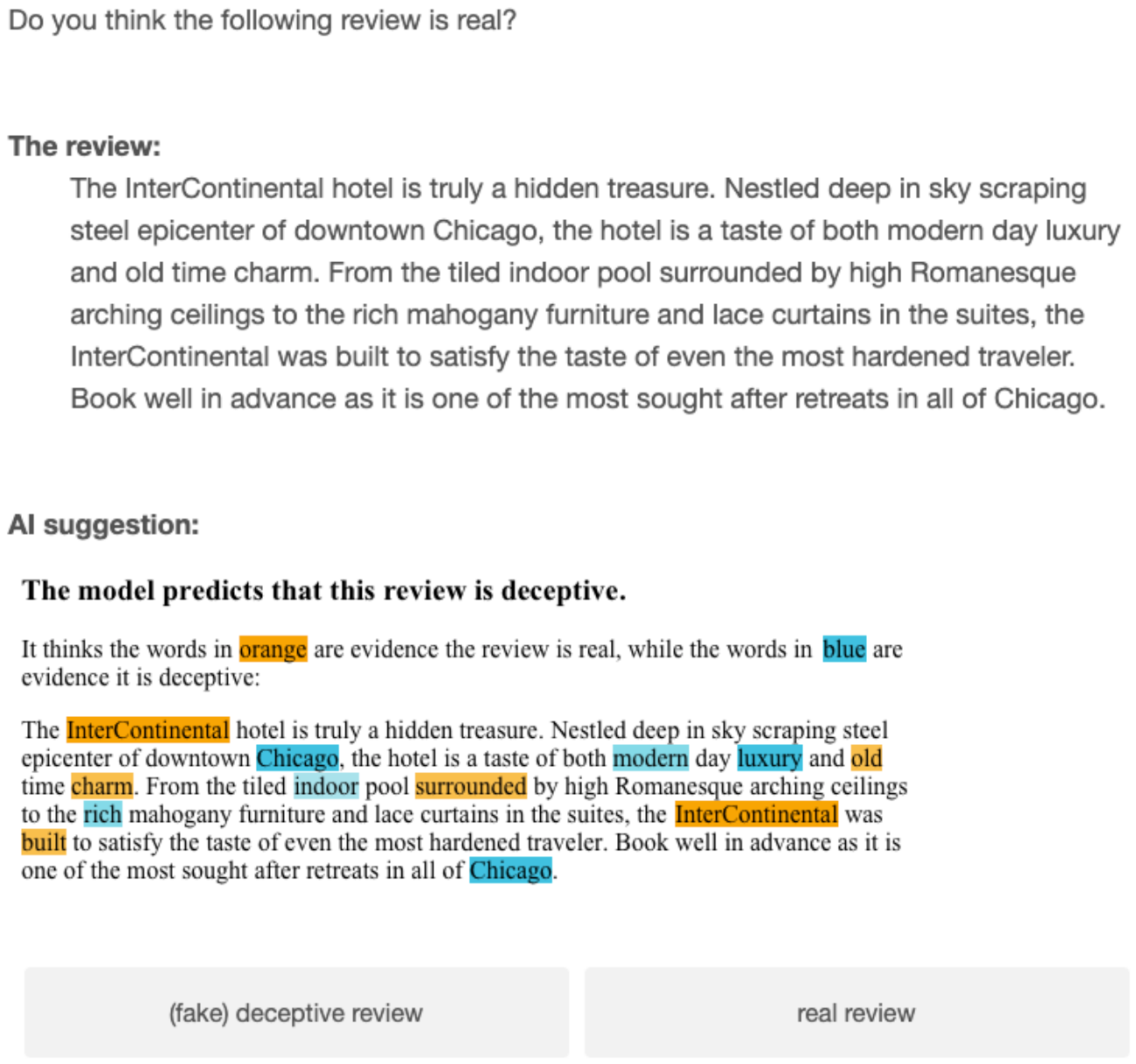}
        \caption{an example question presented to participants in condition $C_3$}
        \label{fig:q3}
    \end{subfigure}
    \caption{One example question from the survey for conditions $C_2$ and $C_3$}
\end{figure}

Each participant was presented the same 20 reviews and AI predictions, and asked to categorize each review as deceptive or real. The only difference between the questions was in the type of AI explanation presented to the participants. We provide an example question from the survey, under each of the 4 conditions. The question and explanation for $C_0$ can be seen in \ref{fig:q0}, the question and explanation for $C_1$ can be seen in \ref{fig:q1}, and the prompts for $C_2$ and $C_3$ are in \ref{fig:q2} and \ref{fig:q3}, respectively.

After being presented with these 20 reviews, respondents were asked to answer questions about their age range and gender identity. Then, they were asked the following three qualitative questions about their experience: \begin{itemize}
    \item \textit{How effective were the AI suggestions in helping you decide if the review was deceptive?} \\ For this question, respondents had the options ``Not effective at all," ``Slightly effective," ``Moderately effective," ``Very effective," and ``Extremely effective."
    \item \textit{How much did you trust the AI suggestions?} \\ For this question, respondents had the options ``Not at all,'' ``A little,'' ``A moderate amount,'' ``A lot,'' and ``A great deal.''
    \item \textit{How difficult did you find this task?} \\ For this question, respondents had the options ``Extremely difficult,'' ``Somewhat difficult,'' ``Neither easy nor difficult,'' ``Somewhat easy,'' and ``Extremely easy.''
\end{itemize}

Finally, participants were asked ``Please explain how (or whether) the AI suggestions affected your decisions.'' They were given a text box to respond as they saw fit.

\subsection{Explanation Validity}
Among the many explanation methods, we use LIME to best adhere to the closed prior work to our endeavor \cite{bansal2021does}. However, since this work does not specifically look at the deceptive reviews task, we run a few validation techniques to ensure that the explanations are meaningful. Note we run this validation on just the base model since lower-quality base model explanations may reduce trust and result in an over-report of the effect in \textsc{H2} (reduction in overreliance) that we are trying to measure.   

\paragraph{Explanation Fidelity}
First, we measure whether LIME outputs for features are consistent with the model output score. For each example $x$ we observe $f(x) \in \{0, 1\}$. We look at the top $15$ features and check whether the sum of the weights of these features amounts to the prediction. For example, an example with all negative features identified by LIME should have the matching $0$ model prediction label. 
Running our validation over the $320$ test examples, only $3\%$ of examples were on the wrong side of the decision threshold ($\tau  = -0.15$ chosen to minimize crossover). And only 1 of these examples was included in our deceptive reviews study. This further suggests that the examples in our study are not particularly difficult to approximate via linear explanations such as LIME. A similar type of analysis could be done for other techniques such as SHAP, but we leave this extension to future work. 

\paragraph{Explanation Agreement (Linear SVM vs LIME)}
\citet{lai2019human} use the actual weights of a linear model for their feature explanations for their human study. To validate that LIME features are sufficiently close to these features, we compute overlap metrics between lime features and our linear SVM features. We observe top feature overlap \textsc{TopK} of 56\% and predictive feature overlap of 60\%. This indicates that there is a significant overlap in top features between Linear SVM and LIME (Compared to results in \cite{krishna2022disagreement}). 

\subsection{Human Baseline}
After running this study, we also ran a human baseline, with 45 participants answering the questions with no model assistance. This baseline was run in exactly the same format as the previous study, except with no model assistance, and correspondingly, the questions related to trust in the model and the instructions related to model assistance removed. The participants in the baseline had an accuracy of $0.61 \pm 0.017$, and a Cohen's Kappa agreement score of $0.115 \pm 0.031$ with the model. The accuracy aligns with the prior work \cite{lai2019human}, which found that humans on their own get an accuracy of around $62\%$ on the Chicago dataset as a whole. Moreover, the Cohen's Kappa agreement score is also expected, as we would expect to have low agreement with a model that the respondents do not see, but a nonzero agreement due to the fact that humans and the model may pick up on similar signs of deceptive reviews. 

The overall accuracy does not significantly change between the human-only condition and any of $C0$ through $C4$ (we get $p$-values between 0.15 and 0.45 when comparing accuracy to each of these conditions). This is perhaps expected, due to the fact that the reference model also gets a $60\%$ accuracy on these questions. However, the answers to these questions do significantly change between conditions; the model agreement for participants in $C1$ (who were presented with one explanation) increases to $0.34 \pm 0.08$, this gives us a $p$-value of $1.1e-5$ when compared with the human-only baseline. In contrast, $C2$ (participants who were presented with dissenting explanations), show a similar agreement to the human-only baseline, and a question-level analysis shows that $C2$ only significantly differs from the baseline on one of the questions; this is expected by the definition of $p$-value for a set of 20 questions where all the answers are drawn from the same distribution.

Thus, these results indicate that humans have a reasonable accuracy on this subset of questions on their own, and though this accuracy does not change, they tend to agree with the model (on both correct and incorrect answers) in conditions where they are presented no or one explanation, and learn to not overrely on the model and follow their own intuition more when presented with dissenting explanations. This is further evidenced by some of the comments from participants in $C2$, such as ``\textit{I didn't feel like most of the keywords the AI was highlighting were really ones that helped me determine real or fake. So I would consult the AI briefly, but then go with my own determination more based on the details given}.''

Based on this baseline, an interesting direction for future study would be to run this experiment on a different task, and with a different form of explanation, so that the model-only accuracy is higher than the human-only accuracy, even on the subset of questions given, and so that the explanations are more human-readable, allowing for further granularity in the difference between the four conditions.

\end{document}